\documentclass[12pt]{article}
\usepackage{graphicx}
\usepackage{enumerate}
\usepackage{url} 
\usepackage{mystylefile}   

\newcommand{\blind}{1}

\begin{document}

\def\spacingset#1{\renewcommand{\baselinestretch}%
{#1}\small\normalsize} \spacingset{1}


\if1\blind
{
  \title{\bf A Projector-Based Approach to Quantifying Total and Excess Uncertainties for Sketched Linear Regression}
  \date{}
  \author{Jocelyn T. Chi
	and 
    Ilse C. F. Ipsen\thanks{
    	The work was supported in part by NSF grants DGE-1633587,
    	DMS-1760374, and DMS-1745654.}
	}
  \maketitle
} \fi

\if0\blind
{
  \bigskip
  \bigskip
  \bigskip
    \title{\bf A Projector-Based Approach to Quantifying Total and Excess Uncertainties for Sketched Linear Regression}
    \date{}
    \author{}
	\maketitle
  \medskip
} \fi

\bigskip
\begin{abstract}
Linear regression is a classic method of data analysis.  In recent years, sketching -- a method of dimension reduction using random sampling, random projections, or both -- has gained popularity as an effective computational approximation when the number of observations greatly exceeds the number of variables.  In this paper, we address the following question: How does sketching affect the statistical properties of the solution and key quantities derived from it?

To answer this question, we present a projector-based approach to sketched linear regression that is exact and that requires minimal assumptions on the sketching matrix.  Therefore, downstream analyses hold exactly and generally for all sketching schemes.  Additionally, a projector-based approach enables derivation of key quantities from classic linear regression that account for the combined model- and algorithm-induced uncertainties.  We demonstrate the usefulness of a projector-based approach in quantifying and enabling insight on excess uncertainties and bias-variance decompositions for sketched linear regression.  Finally, we demonstrate how the insights from our projector-based analyses can be used to produce practical sketching diagnostics to aid the design of judicious sketching schemes.
\end{abstract}

\noindent%
\vfill

\newpage
\spacingset{1.5} 

\section{Introduction}

Linear regression is a classic method of data analysis that is ubiquitous across numerous domains.  In recent years, sketching -- a method of dimension reduction using random sampling, random projections, or a combination of both -- has gained popularity as an effective computational approximation when the number of observations greatly exceeds the number of variables.  In this paper, we address the following question: How does sketching affect the statistical properties of the solution and key statistical quantities derived from it?

To answer this question, we present a projector-based approach to sketched linear regression that is exact and that requires no additional assumptions on the sketching matrix.  Consequently, downstream analyses derived from this formulation of the sketched solution hold exactly and generally for all sketching schemes, while accounting for both model- and algorithmic-induced uncertainties.  


Our paper extends previous work on the combined model- and algorithm-induced uncertainties of the sketched solution to exact expressions that hold generally for \textit{all} sketching schemes.  Specifically, we extend existing work on the total expectation and variance of the sketched solution from specific sampling schemes \cite{MMY14, MMY15} to all sketching schemes.  Due to the assumptions and limitations of a Taylor expansion approach to the solution in \cite{MMY14, MMY15}, the expressions for the total uncertainties there are restricted to specific sampling schemes.  By constrast, our expressions hold for many commonly-used sketching schemes not covered by \cite{MMY14, MMY15}.  These include sketching with fast Fourier Johnston-Lindenstrauss transforms (FJLTs), Gaussian random matrices, and random row-mixing transformations followed by uniform sampling.

We demonstrate the usefulness of a projector-based approach in quantifying and enabling insight on excess uncertainties arising from the randomness in the sketching algorithm.   We highlight this through geometric insights and interpretation for the excess bias and variance, and analyses of total and excess bias-variance decompositions for sketched linear regression.  Finally, we demonstrate how the insights from our projector-based analyses can be used to produce practical sketching diagnostics to aid the design of judicious sketching schemes.

\subsection{Related work}

Randomized sketching is a form of preconditioning and appears to have originated in \cite{Sarlos2006}.  Its many variants can be classified \cite[Section 1]{thanei2017random} according to whether they achieve row compression \cite{BD09,DrineasMM2006,Drineas2011,IpW2014,MMY14,MMY15,RT08,WZM18, RM16}, column compression \cite{AMTol10, thanei2017random, kaban2014new, zhou2008compressed, maillard2009compressed}, or both \cite{LSRN}.  
We focus on \textit{row-sketched linear regression}, where the number of observations greatly exceeds the number of variables.  We refer to this simply as \textit{sketched linear regression}.

Since sketched linear regression has roots in theoretical computer science and numerical analysis, much emphasis has been on analyzing the error due to algorithmic randomization.  Recent works have made progress towards a combined statistical and algorithmic perspective.  These include criteria for quantifying prediction and residual efficiency \cite{RM16}, bootstrap estimates for estimating the combined uncertainty \cite{LWM2018}, approximate expressions for the total expectation and variance of some randomized sampling estimators \cite{MMY14, MMY15}, and asymptotic analysis of randomized sampling estimators \cite{ma2020asymptotic}.

\subsection{Overview}

We present results in terms of two regimes.  The first regime requires no assumptions on the sketching matrix beyond its dimensions.  Consequently, these results hold generally for all sketching matrices and provide a worst-case analysis since they hold even for poor choices of sketching schemes.

The second regime presents results conditioned on rank preservation so that the sketched matrix has the same rank as the original design matrix $\mx$.  Rank preservation implies that the sketching scheme successfully preserves the most relevant information in the original response $\vy$ and design matrix $\mx$.  Although these results require an additional assumption, conditioning on rank preservation enables further insights on how the sketching process affects the solution and other key statistical quantities.  Thus, results from this second regime provide insights from an ideal-case analysis.


\section{Sketched Linear Regression}\label{s_ass}

We begin by setting some notation for the rest of this paper.  We then review the exact and sketched linear regression problems, their solutions, and other relevant quantities.

\subsection{Preliminaries}

Let $\mx\in\real^{n\times p}$ be observed with  $\rank(\mx)=p$.
Since $\mx$ has full column rank, its Moore-Penrose inverse is a left inverse so that
\begin{eqnarray*}\label{e_mp}
\mx^{\dagger}=(\mx^T\mx)^{-1}\mx^T \qquad \text{and} \qquad \mx^{\dagger}\mx = \mi_p.
\end{eqnarray*}
Let $\| \mx \|_{2}$ denote the Euclidean operator norm of $\mx$.    
The two-norm condition number of $\mx$ with regard to left inversion is 
\begin{eqnarray*}
\kappa_2(\mx)\equiv \|\mx\|_2\|\mx^{\dagger}\|_2.
\end{eqnarray*} 
We additionally use $\| \cdot \|_{2}$ to denote the Euclidean vector norm for vectors.  The use of $\| \cdot \|_{2}$ to denote either the Euclidean operator or vector norm will be clear from the context.  Let $\mi_{n}$ denote the $n \times n$ identity matrix, and 
let ${\bf 0}$ and $\myone$ denote the vectors of all zeros and ones, respectively.  Their lengths will be clear from the context.


\subsection{The exact problem and solution}\label{s_lsdet}

Given an observed pair $\vy\in\rn$ and $\mx \in \mathbb{R}^{n \times p}$ with $\rank(\mx) = p$, we assume a Gaussian linear model
\begin{eqnarray}\label{e_g}
\vy=\mx\vbetao+ \veps, \qquad \veps\sim\mathcal{N}(\vzero,\sigma^2 \mi_n),
\end{eqnarray}
where $\vbeta_0\in\real^p$ is the true but unobserved coefficient vector,
and $\veps\in\rn$ is a noise vector with a zero mean multivariate normal distribution and $0 < \sigma^{2} \in \mathbb{R}$.  The unique maximum likelihood estimator of $\vbetao$ is the solution $\vhbeta$ of the exact linear regression problem
\begin{eqnarray}\label{e_d}
\min_{\vbeta\in\real^p}{\|\vy - \mx\vbeta\|_2^2}.
\end{eqnarray}

Since $\mx$ has full column rank, this problem is well posed and has the unique solution
\begin{eqnarray*}\label{e_betahat}
\vhbeta \equiv  \mx^{\dagger}\vy.
\end{eqnarray*}
The exact prediction  and residual are
\begin{eqnarray*}
\vhy \equiv  \mx\vhbeta \qquad \text{and}\qquad
\vhe \equiv \vy -\mx\vhbeta  =  \vy-\vhy,
\end{eqnarray*}
respectively.  The orthogonal projector onto $\range(\mx)$ along $\mynull(\mx^T)$ is 
\begin{eqnarray*}\label{e_hat}
\mPx\equiv  \mx\mx^{\dagger}  =  \mx(\mx^T\mx)^{-1}\mx^T\ \in\rnn
\end{eqnarray*}
and is also known as the \textit{hat matrix} \cite{ChatterH86,HoagW78,VelleW81}.  We express the prediction and residual as 
\begin{eqnarray*}\label{e_haty}
\vhy  = \mPx\vy \qquad  \text{and}\qquad \vhe = (\mi-\mPx)\vy.
\end{eqnarray*}

\subsection{The sketched problem and solution}\label{s_lsrand}

Given an observed matrix-valued random variable $\ms\in\real^{r\times n}$ with $p \le r\leq n$, the sketched linear regression problem
\begin{eqnarray}\label{e_s}
\min_{\vbeta\in\real^p}{\|\ms(\vy - \mx\vbeta)\|_2^2}
\end{eqnarray}
has the minimum norm solution 
\begin{eqnarray*}\label{e_lsrs}
\vtbeta \equiv (\ms\mx)^{\dagger}\,\ms\vy,
\end{eqnarray*}
where $\ms$ is a \textit{sketching matrix}.  
Since we make no assumptions on $\ms$ beyond its dimensions, the sketched matrix $\ms\mx$ may be rank deficient so that \eqref{e_s} may be ill-posed.

By design, $\ms$ has fewer rows than $\mx$. Therefore, the corresponding predictions
$\vhy = \mx\vhbeta$ and $\ms\mx\vtbeta$ have different dimension 
and cannot be directly compared. To remedy this, 
we follow previous work \cite{DrineasMM2006,Drineas2011,RM16}, and compare the predictions
with regard to the \textit{original} design matrix $\mx$.  Therefore, the sketched prediction and residual are
\begin{eqnarray*}\label{e_lsry}
\vty \equiv  \mx\vtbeta \qquad \text{and}\qquad
\vte \equiv  \vy -\mx\vtbeta = \vy-\vty.
\end{eqnarray*}

Sketching can be an effective approach in the highly over-constrained case
\cite{DrineasMM2006, Drineas2011, MMY15, RM16,WZM18, RT08}, where $n$ greatly exceeds $p$.  A standard method of computing the exact solution of \eqref{e_d} is based on a QR decomposition, which requires $\mathcal{O}(n^{2}p)$ operations.  Meanwhile, applying a general sketching matrix requires $\mathcal{O}(rnp)$ operations (fewer when sketching with FJLTs or diagonal sampling matrices) and solving the reduced dimension problem \eqref{e_s} requires $\mathcal{O}(r^{2}p)$ operations.  Thus, computation of a general sketched solution requires $\mathcal{O}(rnp)$ operations so that sketching can offer substantial computational savings for very large $n$ with $r$ significantly smaller than $n$.

\section{A Projector-Based Approach}\label{s_struct}

Given a sketching matrix $\ms$, we view the sketched problem in \eqref{e_s} as a deterministic multiplicative perturbation of the exact problem in \eqref{e_d}.  Therefore, we derive structural bounds for the sketched quantities.  
We begin by presenting an oblique projector for the sketched problem in \eqref{e_s} that plays the role of $\mPx$ in \eqref{e_d}.  This oblique projector enables comparisons between the sketched solution, prediction, and residual and their higher-dimensional exact counterparts.

\begin{lemma}\label{l_proj}
For the sketched problem in \eqref{e_s},
\begin{eqnarray*}
\mP\equiv \mx(\ms\mx)^{\dagger}\ms
\end{eqnarray*}
is an oblique projector where
\begin{eqnarray*}
\mPx\mP   =\mP \quad \text{ and } \quad \mP\mx  = \mx \text{ if } \rank(\ms\mx)=p. 
\end{eqnarray*}
\end{lemma}

These properties follow from the definitions of $\mx\Dag$ and $(\ms\mx)\Dag$.  In general, we have
\begin{eqnarray*}
	\rank(\mP)= \rank(\ms\mx)\le \rank(\mx)=\rank(\mPx)=p
\end{eqnarray*}
so that $\range(\mP) \subseteq \range(\mPx)$.  If $\ms$ preserves rank so that $\rank(\ms\mx)=\rank(\mx)$, then $\range(\mP)=\range(\mPx)$.  However, $\mynull(\mP) = \mynull(\mx\Tra\ms\Tra\ms)$ \cite[Theorem 3.1]{cerny}, so that $\mynull(\mP) \ne \mynull(\mPx)$ in general.  Finally, if $\ms=\mi_n$, then $\mP=\mPx$.

Notice that $\mP$ generalizes $\mPu\equiv \mU(\ms\mU)^{\dagger}\ms$ in \cite[(11)]{RM16}, where $\mU$ is an orthonormal basis for $\range(\mx)$, for quantifying the \textit{prediction efficiency} and \textit{residual efficiency} of sketching algorithms.  However, $\mPu$ is only defined if $\rank(\ms\mx)=\rank(\mx)$ and in that case, $\mPu = \mP$.  Since our analyses extend to $\rank(\ms\mx)<\rank(\mx)$, we employ the more general $\mP$.

Oblique projectors also appear in other contexts.  Examples include constrained least squares \cite{Ste2011, cerny}, weighted least squares \cite{brust2020computationally, MR976336}, discrete inverse problems \cite{MR3021435}, and the discrete empirical interpolation method (DEIM) \cite[Section 3.1]{MR3826673} to name a few.  
We now present the sketched solution, prediction, and residual for \eqref{e_s} in terms of $\mP$. 

\begin{theorem}\label{t_1}
For the sketched problem in \eqref{e_s}, the minimum norm solution is
\begin{eqnarray*}
\vtbeta  =  \mx^{\dagger} \mP\vy = \vhbeta+\mx^{\dagger}(\mP-\mPx)\vy.
\end{eqnarray*}
Therefore, the sketched prediction $\vty=\mx\vtbeta$ and residual $\vte=\vy-\mx\vtbeta$ are
\begin{eqnarray*}
\vty \,=\, \mP\vy \,=\, \vhy+(\mP-\mPx)\vy \quad \text{ and } \quad
 \vte \,= \, \left(\mi-\mP\right)\vy =\vhe+(\mPx-\mP)\vy.
\end{eqnarray*}
\end{theorem}

The expressions for $\vtbeta$, $\vty$, and $\vte$ follow from their definitions in Section \ref{s_ass} and the definitions of $\mP$, $\mP$, $\vhy$, $\vhbeta$, and $\vhe$.  Although the expressions for $\vtbeta$, $\vty$, and $\vte$ in Theorem \ref{t_1} are straightforward, they are exact and hold generally for \textit{all} sketching schemes.  

The significance of Theorem \ref{t_1} is that since it requires no assumptions on $\ms$ (beyond its dimensions) or $\rank(\ms\mx)$, it enables expressions for the total uncertainty due to the combined model- and algorithm-induced randomness for \textit{all} sketching schemes.  These include many commonly-used sketching schemes not covered by previous work \cite{MMY14, MMY15}.  We comparing Theorem \ref{t_1} to a corresponding result in \cite{MMY15}, reproduced below in Lemma \ref{l_mmy}.

\begin{lemma}[Lemma 1 in \cite{MMY15}]\label{l_mmy}
For the sketched problem in \eqref{e_s}, if the following additionally hold: 1) the sketching matrix $\ms$ has a single nonzero entry per row, 2) the vector $\vw\equiv \diag(\ms^T\ms)\in\rn$ has a scaled multinomial distribution with expected value $\E[\vw]=\myone$, 3) $\ms$ preserves rank so that $\rank(\ms\mx)=\rank(\mx)$, and 4) the sketched solution admits a Taylor series expansion around $\E[\vw]$, then
\begin{eqnarray*}
\vtbeta(\vw)=\vhbeta +\mx^{\dagger}\diag(\vhe)(\vw-\myone) + R(\vw),
\end{eqnarray*}
where $R(\vw)$ is the remainder of the Taylor series expansion. 
\end{lemma}

The assumptions in \cite[Lemma 1]{MMY15} and its other versions in \cite{MMY15} limit their scope to sampling schemes where the expected value of the sampling weights vector is known.  Consequently, downtream analysis of the total expectation and variance of the sketched solution using these in \cite{MMY15} are also limited to those same sampling schemes.

Therefore, Theorem \ref{t_1} extends the pioneering work on quantifying the total uncertainties for sketched in linear regression in \cite{MMY14, MMY15} in the following ways.
\begin{enumerate}
	\item First, Theorem \ref{t_1} places no assumptions on $\ms$ or $\rank(\ms\mx)$ so that it applies generally to \textit{all} sketching schemes.  In practice, a wide variety of sketching schemes are used.  These include sketching with fast Johnson-Lindenstrauss transforms (FJLTs), Gaussian transforms, and combinations of FJLTs followed by uniform sampling, to name a few.  Unfortunately, the analysis in \cite{MMY15} does not apply to these.
	\item Second, Theorem \ref{t_1} is exact so that downstream analysis with these expressions do not hinge on the assumptions required for approximations.
	\item Third, framing the sketched solution in terms of the difference between the orthogonal projector $\mPx$ for the exact problem and oblique projector $\mP$ for the sketched problem affords additional geometric insight that we detail later in Sections \ref{s_stat}, \ref{sec:excessbv}, and \ref{sec:bias-var}.
	\item Finally, a projector-based approach greatly simplifies the proofs so that Theorem \ref{t_1} does not require the heavy-duty matrix algebra used to produce the approximate yet more restrictive existing results in \cite{MMY14, MMY15}.
\end{enumerate}

Applying Theorem \ref{t_1} and \cite[(5.3.16)]{GovL13}, which implies that
\begin{eqnarray*}
	\frac{\|\vy\|_2}{\|\mx\|_2\|\vhbeta\|_2} \leq \frac{\|\vy\|_2}{\|\mx\vhbeta\|_2} = \frac{1}{\cos{\theta}},
\end{eqnarray*}
produces the following relative error bounds for the sketched solution and prediction.

\begin{corollary}\label{c_1b}
For the sketched problem in \eqref{e_s}, let
$0<\theta<\frac{\pi}{2}$ be the angle between $\vy$ and $\range(\mx)$. Then the minimum norm sketched solution $\vtbeta$ satisfies 
\begin{eqnarray*}
\frac{\|\vtbeta - \vhbeta\|_2}{\|\vhbeta\|_2} \leq  \kappa_2(\mx) \>\frac{\|\vy\|_2}{\|\mx\|_2\|\vhbeta\|_2}\>\|\mP-\mPx\|_2
\leq \kappa_2(\mx)\>\frac{\|\mP-\mPx\|_2}{\cos{\theta}}.
\end{eqnarray*}
The sketched prediction $\vty=\mx\vtbeta$  satisfies
\begin{eqnarray*}
\frac{\|\vty - \vhy\|_2}{\|\vhy\|_2} &\leq & \frac{\|\mP-\mPx\|_2}{\cos{\theta}}.
\end{eqnarray*}
\end{corollary}

The bounds in Corollary~\ref{c_1b} are tight for $\ms=\mi_n$.
Corollary~\ref{c_1b} implies that the sensitivity of $\vtbeta$ to multiplicative perturbations depends on the deviation of $\mP$ from being an orthogonal projector onto $\range(\mx)$, quantified by $\|\mP-\mPx\|_2$. This distance is amplified, as expected, by the conditioning of $\mx$ with regard to (left) inversion, and by the closeness of $\vy$ to $\range(\mx)$.  Corollary~\ref{c_1b} is an absolute and relative bound since $\|\mPx\|_2=1$.

In contrast to multiplicative perturbation bounds for eigenvalue and singular value
problems \cite{Ips98,Ips99b}, Corollary \ref{c_1b} does not require $\ms$ to be nonsingular or square.
We do not view weighted least squares problems \cite[Section 6.1]{GovL13} as multiplicative perturbations since they employ nonsingular diagonal matrices~$\ms$ for regularization or scaling of discrepancies.

In contrast to additive perturbation bounds (\cite[Section 5.3.6]{GovL13}, \cite[Section 20.1]{Higham2002}, \cite[(3.4)]{Stewart1987}), Corollary \ref{c_1b} requires neither the square of the condition number  nor $\rank(\ms\mx)=\rank(\mx)$.  Therefore, the minimum norm sketched solution $\vtbeta$ and its residual $\vte$ are less sensitive to multiplicative perturbations than to additive perturbations.

Corollary \ref{c_1b} improves on existing structural bounds for sketched least squares algorithms, such as \cite[Theorem 1]{Drineas2011} reproduced in Lemma~\ref{l_cc2} below.

\begin{lemma}[Theorem 1 in \cite{Drineas2011}]\label{l_cc2}
For the sketched problem in \eqref{e_s}, if $\|\mPx\vy\|_2\geq \gamma\, \|\vy\|_2$ for some $0<\gamma\leq 1$
and $\|\vte\|_2\leq (1+\eta)\,\|\vhe\|_2$, then
\begin{eqnarray*}
\frac{\|\vtbeta-\vhbeta\|_2}{\|\vhbeta\|_2}\leq \kappa_2(\mx) \sqrt{\gamma^{-2}-1}\,\sqrt{\eta}.
\end{eqnarray*}
\end{lemma}

Corollary~\ref{c_1b} improves on \cite[Theorem 1]{Drineas2011} in the following ways.  
First, the bound for $\vtbeta$ in Corollary~\ref{c_1b} is more general and tighter as it  does not exhibit nonlinear dependencies on the perturbations.  Second, Corollary~\ref{c_1b} holds under weaker assumptions.  The first inequality for the sketched solution in Corollary~\ref{c_1b} requires only that $\vhbeta \ne {\bf 0}$.  The second inequality for the sketched solution requires only that $\vy \notin \range(\mx)$ and $\vy \notin \range(\mx^{\perp})$.

\section{Model- and Algorithm-Induced Uncertainties}\label{s_stat}
 
The solution $\vhbeta$ of the exact problem in \eqref{e_d} has desirable statistical properties since it is an unbiased estimator of the true coefficient vector $\vbetao$, and it has minimal variance among all linear unbiased estimators of $\vbetao$ (e.g.\@ \cite[Chapter 3, Section 3d]{searle2016linear}).  
A question one might ask is: How does sketching affect the statistical properties of the solution $\vtbeta$ of \eqref{e_s}?


To answer this question, we derive the total expectation and variance due to the combined model- and algorithm-induced uncertainties for the sketched solution $\vtbeta$ and compare them to those of the exact solution $\vhbeta$.  
Since our expressions rely on Theorem \ref{t_1}, our results extend the work in \cite{MMY14, MMY15} to all sketching schemes.

We briefly review the model-induced uncertainty from a Gaussian linear model in Section \ref{s_miu}.  We then derive the expectation and variance of $\vtbeta$ conditioned on the algorithm-induced uncertainty in Section \ref{s_muicond}.  Next, we employ the law of total expectation (e.g.\@ \cite[Theorem 4.4.3]{casella2002statistical}) to derive the total expectation and variance for the combined model- and algorithm-induced uncertainties in Section~\ref{s_total}.  Finally, we visit the total expectation and variance conditioned on sketching schemes that preserve rank in Section \ref{sec:totaluncert-rankpreserved}.  While the latter require an additional assumption, they enable insights that we elaborate on later.

\subsection{Model-induced uncertainty}\label{s_miu}
We refer to the randomness implied by a Gaussian linear model as the \textit{model-induced uncertainty}.  Since the noise vector has mean and variance equal to
\begin{eqnarray*}
	\Ey[\veps] = \vzero \quad \text{ and } \quad \Vy[\veps] = \sigma^2\,\mi_n,
\end{eqnarray*}
the exact solution $\vhbeta$ has mean and variance equal to
\begin{eqnarray}\label{e_miu} 
\Ey[\vhbeta]  =   \vbetao \quad \text{ and } \quad \Vy[\vhbeta] = \sigma^2(\mx^T\mx)^{-1}\in\real^{p\times p}.
\end{eqnarray}
It is well-known that the variance of $\vhbeta$ depends on the conditioning of~$\mx$ \cite[Section 5]{Stewart1987}.  

A difficulty in analyzing row-sketching \eqref{e_s}, coupled with general concern regarding first-order expansions like the ones in \cite{MMY14,MMY15}, is potential rank deficiency in the sketched matrix so that $\rank(\ms\mx)<\rank(\mx)$.  In this case, $(\ms\mx)^{\dagger}$ cannot be expressed in terms of $\ms\mx$.  Thus, we introduce a projector that quantifies the bias arising from rank deficiency in $\ms\mx$. 

\begin{lemma}[Bias projector]\label{l_bias}
For the sketched problem in \eqref{e_s}, 
\begin{eqnarray*}
\mPo \equiv (\ms\mx)^{\dagger}(\ms\mx)\in\real^{p\times p}
\end{eqnarray*}
is an orthogonal projector with the following consequences
\begin{eqnarray*}
	\mP\mx=\mx\mPo \quad \text{ and } \quad \mPo=\mi_p \text{ if } \rank(\ms\mx)=p. 
\end{eqnarray*}
\end{lemma}

Orthogonality follows from $(\mPo)^2=\mPo$ and $(\mPo)^T=\mPo$, which follow from the fact that $(\ms\mx)\Dag$ is a Moore-Penrose generalized inverse.  
If $\rank(\ms\mx)<p$, then $\mPo$ characterizes the subspace of $\range(\mx)$ onto which~$\mP$ projects.
The name \textit{bias projector} will become apparent in Theorem~\ref{l_12}, where $\mPo$ quantifies the bias in $\vtbeta$.

\subsection{Conditional expectation and variance}\label{s_muicond}

We condition on a given sketching matrix $\ms$ and derive the conditional model-induced expectation and variance of the sketched solution $\vtbeta$.  Theorem \ref{l_12} below shows that the conditional expectation depends on the bias projector $\mPo$ while the conditional variance depends on the oblique projector $\mP$.

\begin{theorem}[Model-induced uncertainty conditioned on $\ms$]\label{l_12}
For the sketched problem in \eqref{e_s}, the solution $\vtbeta$ has conditional expectation
\begin{eqnarray*}
\Ey[\vtbeta \,|\, \ms] \;=\; \mPo\vbetao \;=\; \vbetao -(\mi-\mPo)\vbetao,
\end{eqnarray*}
where $\mi-\mPo$ quantifies the rank deficiency of $\ms\mx$, 
and conditional variance
\begin{eqnarray*}
\Vy[\vtbeta \,|\, \ms] & = & \sigma^2 \,\left(\mx^{\dagger}\mP\right)\left(\mx^{\dagger}\mP\right)^T\\
&=& \Vy[\vhbeta] +\sigma^2\,\mx^{\dagger}\left(\mP\mP^T-\mPx\right)(\mx^{\dagger})^T,
\end{eqnarray*}
where $\mP\mP^T-\mPx$ represents the deviation of $\mP$ from being an orthogonal projector onto $\range(\mx)$.
\end{theorem}

\begin{proof}
For the conditional expectation, we employ the second expression for $\vtbeta$ in Theorem \ref{t_1}.  The result follows from the fact that $\mx\Dag$ is a left inverse for $\mx$ and the definition of $\mPo$.

For the first expression for the conditional variance, we apply the definition of the variance conditioned on $\ms$ to the first expression for $\vtbeta$ in Theorem \ref{t_1}.  We combine this with the expression for the conditional expectation for $\vtbeta$ to obtain
\begin{eqnarray}
	\Vy[\vtbeta \,|\, \ms] &=& \Ey[\vtbeta\vtbeta^T\, |\, \ms] -
	\Ey[\vtbeta\,|\, \ms] \,\Ey[\vtbeta\,|\, \ms]^T \notag\\
	&=& \left(\mx^{\dagger}\mP\right) \Ey[\vy\vy^T] \left(\mx^{\dagger}\mP\right)^T- (\mPo\vbetao)(\mPo\vbetao)^T. \label{eq:condvar1}
\end{eqnarray}
Expanding the middle term in the first summand gives
\begin{eqnarray}\label{e_Eyy}
\Ey[\vy\vy^T]&=&(\mx\vbetao)(\mx\vbetao)^T 
+\Ey[\veps\veps^T]\nonumber\\
&=&(\mx\vbetao)(\mx\vbetao)^T+\sigma^2\mi_n.
\end{eqnarray}
We then substitute \eqref{e_Eyy} into \eqref{eq:condvar1}.  Using the fact that $\mx\Dag \mP \mx = \mPo$ and canceling terms produces the first expression.  
For the second expression for the conditional variance, we use the facts that 
\begin{eqnarray*}
	\mx\Dag \mPx = \mx\Dag \quad \text{ and } \quad
	\mx\Dag(\mx\Dag)^T = (\mx\Tra\mx)\Inv
\end{eqnarray*}
to rewrite $\Vy[\vhbeta]$ in \eqref{e_miu} as
\begin{eqnarray}
	\Vy[\vhbeta]=\sigma^2\,\mx^{\dagger} \mPx(\mx^{\dagger})^T. \label{eq:varbetahat2}
\end{eqnarray}
The result follows from adding and subtracting \eqref{eq:varbetahat2} in the first expression for the conditional variance. 

For the interpretation of $\mi - \mPo$, notice that if $\ms\mx$ has full column rank, then $\mPo = \mi$.  Therefore, $\mi - \mPo$ represents the deviation of $\ms\mx$ from having full column rank.  

For the interpretation of $\mP\mP^T  - \mPx$, notice that since $\range(\mP) \subseteq \range(\mPx)$, $\mP$ projects onto a subspace of $\range(\mx)$.  If additionally, $\mP$ is an orthogonal projector, symmetry requires $\ms = \mi_{n}$ so that $\mP=\mP\mP^T=\mPx$.  Therefore, $\mP\mP^T-\mPx$ represents the deviation of $\mP$ from being an orthogonal projector onto $\range(\mx)$.
\end{proof}


Theorem \ref{l_12} shows that the conditional expectation of $\vtbeta$ depends on the rank deficiency of $\ms\mx$.  In particular, the conditional bias of~$\vtbeta$ is proportional to  the deviation $\mi-\mPo$ of $\ms\mx$ from having full column rank.  
To see this, notice that conditioned on $\ms\mx$ having full column rank, $\mPo = \mi$.  In this case, $\mi - \mPo$ vanishes and $\vtbeta$ is a conditionally unbiased estimator of~$\vbetao$ with 
\begin{eqnarray*}
	\Ey[\vtbeta \,|\, \rank(\ms\mx)=\rank(\mx)] = \vbetao.
\end{eqnarray*}  
Since this holds for any $\ms$, the conditional bias of $\vtbeta$ depends only on $\rank(\ms\mx)$.

Theorem \ref{l_12} also shows that the conditional variance of $\vtbeta$ depends on the deviation of $\mP$ from being an orthogonal projector onto $\range(\mx)$. In particular, the conditional variance $\Vy[\vtbeta \,|\, \ms]$ is close to the model variance $\Vy[\vhbeta]$ if $\mP$ is close to $\mPx$.  In the extreme case that $\ms=\mi_n$, the conditional variance is identical to the model variance.  
Corollary \ref{c_l12a} follows directly from Theorem \ref{l_12} and further highlights the relevance of $\mi-\mPo$ and $\mP\mP^T-\mPx$.

\begin{corollary}[Relative differences between conditional and model uncertainties]
	\label{c_l12a}
Given the assumptions in Theorem~\ref{l_12}, we have
\begin{eqnarray*}
\|\Ey[\vtbeta \, |\, \ms] - \vbetao\|_2 &\leq& \|\mi-\mPo\|_2 \, \|\vbetao\|_2
\end{eqnarray*}
and
\begin{eqnarray*}
\frac{\|\Vy[\vtbeta \, |\, \ms] - \Vy[\vhbeta] \|_2}{\|\Vy[\vhbeta]\|_2} &\leq &\|\mP\mP^T-\mPx\|_2.
\end{eqnarray*}
\end{corollary}

The relative conditional variance follows from Theorem \ref{l_12} and the facts that $\|\mx\Dag\|_2\,\|(\mx\Dag)^T\|_2=\|\mx\Dag(\mx\Dag)^T\|_2$,  $\mx\Dag(\mx\Dag)^T = (\mx^T\mx)\Inv$, and $\sigma^{2} > 0$ so that $\|\Vy[\vhbeta]\|_{2} \ne 0$.

Corollary \ref{c_l12a} shows that the relative differences in the conditional bias and variance can be expressed solely in terms of $\mi-\mPo$ and $\mP\mP^T-\mPx$.
In particular, the conditional bias of $\vtbeta$ increases with rank deficiency in $\ms\mx$.  Additionally, the relative difference between conditional and model variances increases with the deviation of $\mP$ from $\mPx$.

Therefore, Corollary \ref{c_l12a} shows that unbiasedness is more readily achievable since it requires only that $\ms\mx$ have full column
rank.  Meanwhile, the conditional variance of $\vtbeta$ is guaranteed to be at least as large as $\Vy[\vhbeta]$, with equality only when $\ms = \mi_{n}$ so that $\mP = \mPx$.  In this case, the sketched problem in \eqref{e_s} becomes the exact problem in \eqref{e_d}.  

\subsection{Total expectation and variance}\label{s_total}

We now view the sketching matrix $\ms$ as a matrix-valued random variable and derive the total expectation and variance of the sketched solution $\vtbeta$.  We employ the expressions for the conditional expectation and variance in Section \ref{s_muicond} and the law of total expectation.

\begin{theorem}[Total uncertainty]\label{t_11}
For the sketched problem in \eqref{e_s}, the solution $\vtbeta$ has total expectation
\begin{eqnarray*}
\E[\vtbeta] &  =   & 
				\vbetao-\left(\mi - \Es[\mPo]\right)\vbetao
\end{eqnarray*}
and total variance
\begin{eqnarray*}
\V[\vtbeta] 
&=&\V[\vhbeta]+\sigma^2\,\mx\Dag\left(\Es[\mP\mP^T]-\mPx\right)(\mx\Dag)^T
+ \Vs[\mPo\vbetao].
\end{eqnarray*}
\end{theorem}

\begin{proof}
For the total expectation, we combine our expression for $\Ey[\vtbeta \,|\, \ms]$ from Theorem \ref{l_12} with the law of total expectation.  For the total variance, we apply the expression for the total expectation in the definition of the variance to obtain
\begin{eqnarray}
	\V[\vtbeta] &=& \E[\vtbeta\vtbeta^T]- \E[\vtbeta]\E[\vtbeta]^T \notag\\
	&=& \Es\left[\Ey\left[\vtbeta\vtbeta^T\, \Big|\,  \ms\right]\right] -
	\left(\Es[\mPo]\vbetao\right)\left(\Es[\mPo]\vbetao\right)^T. \label{eq:var1}
\end{eqnarray}
From \eqref{eq:condvar1} and \eqref{e_Eyy}, we have
\begin{eqnarray}
	\Ey\left[\vtbeta\vtbeta^T\, \Big|\,  \ms\right]
	=\sigma^2 \mx^{\dagger}\mP\mP^T(\mx^{\dagger})^T+(\mPo\vbetao)(\mPo\vbetao)^T. \label{eq:var2}
\end{eqnarray}
Inserting \eqref{eq:var2} into \eqref{eq:var1} then gives us
\begin{eqnarray*}
	\V[\vtbeta]&=& \sigma^2 \mx^{\dagger}\Es\left[\mP\mP^T\right](\mx^{\dagger})^T\\
	&&\quad +\underbrace{\Es\left[\left(\mPo\vbeta_0\right)\left(\mPo\vbeta_0\right)^T\right]
		-\left(\Es[\mPo]\vbetao\right)\left(\Es[\mPo]\vbetao\right)^T}_{\Vs[\mPo\vbetao]},
\end{eqnarray*}
where the latter two terms in the above expression are equal to $\Vs[\mPo\vbetao]$.  Finally, using the fact that $\mx\Dag\mPx(\mx\Dag)\Tra = (\mx\Tra\mx)\Inv$, we add and subtract $\V[\vhbeta]$ from the above expression to obtain the result. 
\end{proof}

Theorem \ref{t_11} shows that the total bias of $\vtbeta$ is proportional to the expected deviation of the matrix-valued random variable $\ms\mx$ from having full column rank.  Therefore, after accounting for both the model- and algorithm-induced uncertainties, the bias of $\vtbeta$ depends on the expected value of $\mPo$.  Notice, however, that the expectation $\Es[\mPo]$ of a projector $\mPo$ is not a projector in general.  

Theorem \ref{t_11} also shows that the total variance of $\vtbeta$ can be decomposed into the following three components:
\begin{enumerate}
	\item the inherent model variance in $\vhbeta$,
	\item the expected deviation of the matrix-valued random variable $\mP$ from being an orthogonal projector onto $\range(\mx)$, and 
	\item the variance in the rank deficiency of the matrix-valued random variable $\ms\mx$ as captured through the bias projector $\mPo$.
\end{enumerate}

Corollary \ref{c_11} follows from Theorem \ref{t_11}.  It shows how rank deficiency, as quantified by $\mi - \mPo$, and the deviation of $\mP$ from being an orthogonal projector, as quantified by $\mP\mP^T - \mPx$, affect the relative differences between the total and model uncertainties.

\begin{corollary}[Relative differences between total and model uncertainties]\label{c_11}
	Given the assumptions in Theorem \ref{t_11}, we have
	\begin{eqnarray*}
		\|\E[\vtbeta] -\vbetao\|_2&  \leq   & \|\mi-\Es[\mPo]\|_2\,\|\vbetao\|_2
	\end{eqnarray*}
	and
	\begin{eqnarray*}
		\frac{\|\V[\vtbeta] -\Vy[\vhbeta]\|_2}{\|\Vy[\vhbeta]\|_2} &\leq &\|\Es[\mP\mP^T]-\mPx\|_2
		+ \frac{\|\Vs[(\mi-\mPo)\vbetao]\|_2}{\|\Vy[\vhbeta]\|_2}.
	\end{eqnarray*}
\end{corollary}

Compared with Corollary \ref{c_l12a}, where the difference between the conditional and model variance depends only on $\mP\mP^T - \mPx$, Corollary \ref{c_11} shows that the difference between the total and model variance depends on two sources.  The first is the expected deviation of $\mP$ from being an orthogonal projector as quantified in $\Es[\mP\mP^T] - \mPx$.  The second is the ratio of the variance of the estimation distortion due to rank deficiency to the model variance.  If the variance in the distortion due to rank deficiency is small relative to the model variance, then this latter term is likewise small.  

\subsection{Total uncertainties conditioned on rank preservation}\label{sec:totaluncert-rankpreserved}

In the previous sections, we worked towards deriving unconditional expressions quantifying the combined model- and algorithm-induced uncertainties in sketched linear regression.  Since those expressions require no assumptions on the sketching matrix $\ms$ beyond its dimensions, they hold exactly and in general for all sketching schemes.

We now present results that condition on sketching matrices that preserve rank so that $\rank(\ms\mx) = \rank(\mx)$.  Although these results require an additional assumption, conditioning on rank preservation enables further insight, which we detail below and in other following sections.  

\begin{corollary}[Total uncertainty conditioned on rank preservation] \label{cor:totaluncertainty-rankpreserved}
	For the sketched problem in \eqref{e_s} conditioned on $\rank(\ms\mx) = \rank(\mx)$, the solution $\vtbeta$ has total expectation
	\begin{eqnarray*}
		\E[\vtbeta] &  =   & \vbetao
	\end{eqnarray*}
	and total variance
	\begin{eqnarray*}
		\V[\vtbeta] 
		&=&\V[\vhbeta]+\sigma^2\,\mx\Dag\left(\Es[\mP\mP^T]-\mPx\right)(\mx\Dag)^T.
	\end{eqnarray*}
\end{corollary}

The expressions for the total expectation and variance follow from Theorem \ref{t_11} and the fact that
$\Es[\mPo \, | \, \rank(\ms\mx) = \rank(\mx)] = \mi$.
Corollary \ref{cor:totaluncertainty-rankpreserved} shows that conditioning on rank preservation, the sketched solution $\vtbeta$ is an unbiased estimator of $\vbetao$.  Later in Corollary \ref{cor:nonnegativity}, we will find that even in these cases, however, the total variance of $\vtbeta$ is at least as great as the model variance $\V[\vhbeta]$.

Compared with \cite[Lemma 2]{MMY15} which also assumes rank preservation, Corollary \ref{cor:totaluncertainty-rankpreserved} is more general in that it holds for all sketching matrices, without restriction to specific kinds of sampling matrices.  Additionally, \cite[Lemma 2]{MMY15}, has an additional term due to the variance of the Taylor expansion remainder.  Corollary \ref{cor:totaluncertainty-rankpreserved} lacks this term since the projector-based formulation of the $\vtbeta$ in Theorem \ref{t_1} holds exactly without any additional assumptions.

\section{Total Excess Bias and Variance}\label{sec:excessbv}

We summarize and interpret the \textit{excess bias} and \textit{excess variance} attributable to algorithm-induced uncertainties.  These represent the additional bias and variance in the sketched solution $\vtbeta$ beyond the model bias $\Bias(\vhbeta, \vbetao)$ and model variance $\Var(\vhbeta)$ arising from the assumptions of a Gaussian linear model.   
We show that the projector-based approach in Theorem \ref{t_1} enables insight and understanding into the sources of excess bias and variance.

\begin{corollary}[Total excess bias and variance]\label{cor:totexcessbv}
	For the problem in \eqref{e_s}, the solution $\vtbeta$ has total excess bias equal to
	\begin{eqnarray*}
		\mathcal{B} &\equiv& (\Es[\mPo] - \mi) \vbetao
	\end{eqnarray*} 
	and total excess variance equal to
	\begin{eqnarray*}
		\mathcal{V} &\equiv& \underbrace{\sigma^2\,\mx\Dag\left(\Es[\mP\mP^T]-\mPx\right)(\mx\Dag)^T}_{\mathcal{V}_{\mP}}
		+ \underbrace{\Vs[\mPo\vbetao]}_{\mathcal{V}_{\mPo}}.
	\end{eqnarray*}
\end{corollary}

Corollary \ref{cor:totexcessbv} follows from Theorem \ref{t_11} and the fact that the exact solution $\vhbeta$ is an unbiased estimator of $\vbetao$.  Recall that $\Es[\mPo] - \mi$ represents the expected deviation of the sketched matrix $\ms\mx$ from having full column rank.  Therefore, the \textit{excess bias} $\mathcal{B}$ represents the expected estimation distortion under rank deficiency from sketching.

Corollary \ref{cor:totexcessbv} shows that we can decompose the \textit{excess variance} $\mathcal{V}$ due to randomness in the sketching algorithm into two sources.  The first source $\mathcal{V}_{\mP}$ is due to the expected deviation of the oblique projector $\mP$ from being an orthogonal projector onto $\range(\mx)$.  The second source $\mathcal{V}_{\mPo}$ arises from the variance of the estimation distortion under rank deficiency from sketching.  
Conditioning on rank preservation so that $\rank(\ms\mx) = \rank(\mx)$ presents simplifications that enable additional insights on the total excess bias and variance.

\begin{corollary}[Total excess bias and variance conditioned on rank preservation]\label{cor:totexcessbv-rankpreserved}
	For the problem in \eqref{e_s} conditioned on $\rank(\ms\mx) = \rank(\mx)$, the solution $\vtbeta$ has zero total excess bias and total excess variance equal to
	\begin{eqnarray*}
		\mathcal{V}' &\equiv& \underbrace{\sigma^2\,\mx\Dag\left(\Es[\mP\mP^T]-\mPx\right)(\mx\Dag)^T}_{\mathcal{V}_{\mP}}.
	\end{eqnarray*}
\end{corollary}

Corollary \ref{cor:totexcessbv-rankpreserved} follows from Corollary \ref{cor:totaluncertainty-rankpreserved}.  Conditioning on rank preservation, both the excess bias $\mathcal{B}$ and the excess variance due to rank deficiency $\mathcal{V}_{\mPo}$ vanish.  Therefore, the excess variance conditioned on rank preservation $\mathcal{V}'$ is equal to $\mathcal{V}_{\mP}$, which quantifies the excess variance arising from the expected deviation of $\mP$ from $\mPx$.

For further interpretation of $\mathcal{V}_{\mPx}$, we revisit the range and null spaces of $\mP$ and $\mPx$.  Recall that if $\rank(\ms\mx) = \rank(\mx)$, we have
\begin{eqnarray*}
	\range(\mP) = \range(\mPx).
\end{eqnarray*}
The fact that $\range(\mP) \subseteq \range(\mPx)$ follows from the identity $\mPx \mP = \mP$.  Additionlly, the fact that $\range(\mPx) \subseteq \range(\mP)$ follows from the identity $\mP \mPx = \mPx$.  Equality therefore follows from double containment.  
Meanwhile, from \cite[Theorem 3.1]{cerny} we have
\begin{eqnarray*}
	\mynull(\mP) = \mynull(\mx\Tra\ms\Tra\ms) \ne \mynull(\mx\Tra) = \mynull(\mPx)
\end{eqnarray*}
in general.  
Thus, we observe how sketching perturbs the subspaces from the exact problem.  If $\rank(\ms\mx) = \rank(\mx)$, the sketching and orthogonal projectors, $\mP$ and $\mPx$, have the same range.  However, the dimension reduction achieved through sketching comes at the cost of a perturbation of $\mynull(\mPx)$.

Therefore, the excess variance arising from the deviation of $\mP$ from $\mPx$ reflects the perturbation of the original subspaces due to algorithm-induced randomness.  Specifically, the deviation of $\mP$ from $\mPx$ in $\mathcal{V}_{\mPx}$ conditioned on rank preservation reflects the deviation of $\mynull(\mP)$ from $\mynull(\mPx)$.


\begin{corollary}[Non-negativity of the total excess variance conditioned on rank preservation]\label{cor:nonnegativity}
	For the problem in \eqref{e_s} conditioned on $\rank(\ms\mx) = \rank(\mx)$, we have
	\begin{eqnarray*}
		\V[\vtbeta] \succcurlyeq \V[\vhbeta],
	\end{eqnarray*}
	where the $\succcurlyeq$ operator denotes the Loewner ordering for symmetric matrices of the same dimension.
	Additionally, we have
	\begin{eqnarray*}
		\trace(\mathcal{V}_{\mP}) \ge 0 \quad \text{ so that } \quad \trace(\Var[\vtbeta]) \ge \trace(\Var[\vhbeta]).
	\end{eqnarray*}
\end{corollary}

\begin{proof}
Corollary \ref{cor:nonnegativity} follows from the fact that conditioning on rank preservation gives the identity $\mP \mPx \mP\Tra  = \mPx$.  Therefore, $\mathcal{V}_{\mP}$ is positive semi-definite since $\mi - \mPx$ is idempotent.  The variance inequalities follow from the fact that positive semi-definite matrices have non-negative trace.
\end{proof}

The facts that $\V[\vtbeta] \succcurlyeq \V[\vhbeta]$ and $\trace(\Var[\vtbeta]) \ge \trace(\Var[\vhbeta])$ are unsurprising in themselves since $\vhbeta$ is the best linear unbiased estimator of $\vbetao$ (e.g. \cite[Chapter 3, Section 3d]{searle2016linear}).  What is surprising, however, is that the projector-based approach shows directly that the additional variance is due to the expected deviation of $\mynull(\mP)$ from $\mynull(\mPx)$.



\section{Bias-Variance Decompositions}\label{sec:bias-var}

We show that the projector-based approach combined with the total uncertainty quantities from Section \ref{s_total} further enable bias-variance decompositions that hold generally for all sketching schemes.  We begin by analyzing the mean squared error for the true parameter $\vbetao$.  We then examine the predictive risk, which in this case is the mean squared error for the true prediction $\mx\vbetao$.  We employ the $\MSE(\cdot,\cdot)$ and $\Risk(\cdot, \cdot)$ operators to denote the mean squared error and predictive risk between two vectors of the same dimension, respectively.

\begin{corollary}[Total mean squared error]\label{cor:mse}
	For the problem in \eqref{e_s}, the solution $\vtbeta$ has total mean squared error equal to
	\begin{eqnarray*}
		\MSE(\vtbeta, \vbetao) &=& \trace\{ \V[\vhbeta]\}+ \sigma^2\,\trace\{\mx^{\dagger}\left(\Es[\mP\mP^T]-\mPx\right)(\mx^{\dagger})^T\} \\ 
		& \;\;\;\;\;& + \trace\{\Vs[\mPo\vbetao]\}
		 + \|\left(\mi - \Es[\mPo]\right)\vbetao\|^{2}_{2}.
	\end{eqnarray*}
\end{corollary} 

\begin{proof}
	We employ the properties of the trace operator and linearity of the trace and expectation to obtain the well-known bias-variance trade-off in terms of the trace operator
	\begin{eqnarray*}
		\MSE(\betatilde, \betanot) &=& \E[\lVert \betatilde - \betanot \rVert^{2}_{2}] \\
		&=& \E[\|\vtbeta - \E[\vtbeta]\|_{2}^{2}] + \| \E[\vtbeta] - \vbetao\|_{2}^{2} \\
		&=& \trace\{\V[\vtbeta]\} + \| \Bias(\vtbeta,\vbetao)\|^{2}_{2}.
	\end{eqnarray*}
	The result follows directly from applying the expressions for the total variance and bias of $\vtbeta$ from Theorem \ref{t_11}.
\end{proof}

Corollary \ref{cor:mse} directly states how the bias and variance of $\vtbeta$ contribute to the total mean squared error.  Specifically, the portion of the total mean squared error due to variance includes the following: 1) $\trace\{\Var[\vhbeta]\}$ -- the variance due to randomness from the model assumptions; 2) $\sigma^2\,\trace\{\mx^{\dagger}\left(\Es[\mP\mP^T]-\mPx\right)(\mx^{\dagger})^T\}$ -- the excess variance due to the deviation of the oblique projector $\mP$ from being an orthogonal projector onto $\range(\mx)$; and 3) $\trace\{\Vs[\mPo\vbetao]\}$ -- the excess variance due to rank deficiency arising from randomness in the sketching algorithm.  
Additionally, the bias portion of the total mean squared error represents the excess bias due to rank deficiency from the sketching process.

The \textit{total excess mean squared error} denotes the portion of the mean squared error attributable to randomness in the sketching algorithm.  This represents the portion of $\MSE(\vtbeta, \vbetao)$ exceeding $\MSE(\vhbeta, \vbetao)$, the mean squared error due to model-induced randomness.  Using the notation in Section \ref{sec:excessbv}, we can rewrite the total mean squared error for the sketched solution $\vtbeta$ as
\begin{eqnarray*}
	\MSE(\vtbeta, \vbetao) &=& \MSE(\vhbeta, \vbetao) + \underbrace{\trace\{ \mathcal{V}_{\mP}\} +  \trace\{ \mathcal{V}_{\mPo}\} + \| \mathcal{B} \|^{2}_{2}}_{\mathcal{M}},
\end{eqnarray*}
where $\mathcal{M}$ denotes the \textit{total excess mean squared error}.  Thus, the excess total mean squared error can be decomposed into three sources with interpretation as stated above.  Conditioning on sketching schemes that preserve rank provides simplifications and additional insights on the total mean squared error.

\begin{corollary}[Total mean squared error conditioned on rank preservation] \label{cor:mse-rankpreserved}
	For the problem in \eqref{e_s} conditioned on $\rank(\ms\mx) = \rank(\mx)$, the solution $\vtbeta$ has total mean squared error
	\begin{eqnarray*}
		\MSE(\vtbeta, \vbetao) & = 
		\trace\{\V[\vhbeta]\}  + \sigma^{2}\trace\{\mx\Dag\left(\Es[\mP\mP^T]-\mPx\right)(\mx\Dag)^T\}.
	\end{eqnarray*}
	Therefore, we additionally have
	\begin{eqnarray*}
		\MSE(\vtbeta, \vbetao) \ge \MSE(\vhbeta, \vbetao).
	\end{eqnarray*}
\end{corollary}

\begin{proof}
	The expression for the mean squared error follows from the fact that both $\vtbeta$ and $\vhbeta$ are unbiased estimators of $\vbetao$ in this case.  Therefore, the mean squared error is the trace of the variance.  For the inequality, we again employ the properties of the trace operator and linearity of the trace and expectation to obtain
	\begin{eqnarray*}
		\MSE(\betatilde, \betanot) &=& \E[\lVert \betatilde - \betanot \rVert^{2}_{2}] \\
		&=& \trace \{ \E[ (\betatilde - \betanot) (\betatilde - \betanot)^T] \}
		= \trace \{ \Var(\betatilde) \} \\
		&=& \sigma^{2}\trace \{ \XtXInv \}  + \sigma^{2}\trace\{\mx\Dag\left(\Es[\mP\mP^T]-\mPx\right)(\mx\Dag)^T\} \\
		&\ge& \MSE(\vhbeta, \betanot).
	\end{eqnarray*}
	Once again, conditioning on rank preservation gives us $\mP\mPx\mP\Tra = \mPx$ so that $\mathcal{V}_{\mP}$ is positive semi-definite since $\mi - \mPx$ is idempotent.  
	Since the trace of a positive semi-definite matrix is non-negative, the result follows from the fact that $\vhbeta$ is an unbiased estimator of $\vbetao$. 
\end{proof}

Corollary \ref{cor:mse-rankpreserved} shows that when conditioning on rank preservation, the excess bias and variance due to rank deficiency, $\mathcal{B}$ and $\mathcal{V}_{\mPo}$, vanish.  Therefore, the \textit{excess total mean squared error} in this case is simply
\begin{eqnarray*}
	\mathcal{M}' &\equiv& \sigma^{2}\trace\{\mx\Dag\left(\Es[\mP\mP^T]-\mPx\right)(\mx\Dag)^T\} = \trace\{\mathcal{V}_{\mPx}\}.
\end{eqnarray*}
As we saw in the explanation of $\mathcal{V}_{\mPx}$ following Corollary \ref{cor:totexcessbv-rankpreserved}, $\mathcal{V}_{\mPx}$ in this case quantifies the excess variance due to the deviation of $\mynull(\mP)$ from $\mynull(\mPx)$.

Corollary \ref{cor:mse-rankpreserved} also shows that even conditioning on rank preservation so that $\vtbeta$ is an unbiased estimator of $\vbetao$, the total mean squared error of $\vtbeta$ is at least as great as that of $\vhbeta$.  The decomposition of the total mean squared error in Corollary \ref{cor:mse-rankpreserved} shows that there are two reasons for this.  First, $\vtbeta$ inherits the model variance $\V(\vhbeta)$.  Second, $\vtbeta$ additionally acquires excess variance $\mathcal{V}_{\mPx}$ from the perturbation of $\mynull(\mPx)$ through sketching.

\begin{corollary}[Total predictive risk]\label{cor:predictiverisk}
	For the problem in \eqref{e_s}, the solution $\vtbeta$ has total predictive risk equal to
	\begin{eqnarray*}
		\Risk(\vty, \mx\vbetao) &=& \Risk(\vhy, \mx\vbetao) + \sigma^2\trace\{\Es[\mP\mP\Tra] - \mPx\} \\
		&\;\;\;\;\;& + \|(\Es[\mP\mP\Tra] - \mPx)\mx\vbetao\|_{2}^{2}.
	\end{eqnarray*}
\end{corollary}

\begin{proof}
	Using the properties of the trace operator and the linearity of the trace and expectation, we obtain the following bias-variance decomposition for the predictive risk
	\begin{eqnarray*}
		\Risk(\vty, \mx\vbetao) &=& \E[\|\vty-\mx\vbetao\|^{2}_{2}] 
		= \trace\{\Var[\vty]\} + \|\Bias(\vty, \mx\vbetao)\|^{2}_{2}.
	\end{eqnarray*}	
	The total variance of $\vty$ follows from applying the law of total expectation to the sketched prediction $\mP\vy$.  The result follows from the facts that $\vhy$ is an unbiased estimator for $\mx\vbetao$ so that $\Risk(\vhy, \mx\vbetao) = \Var[\vhy]$ and $\mPx\mx = \mx$.
\end{proof}

Corollary \ref{cor:predictiverisk} shows that the predictive risk can be decomposed into the following three  sources: 1) $\Risk(\vhy, \mx\vbetao)$ -- the prediction variance inherent in the model; 2) $\sigma^2\trace\{\Es[\mP\mP\Tra] - \mPx\}$ -- the excess prediction variance due to the expected deviation of $\mP$ from $\mPx$; and 3) $\|(\Es[\mP\mP\Tra] - \mPx)\mx\vbetao\|_{2}^{2}$ -- the excess prediction bias arising from the expected deviation of $\mP$ from $\mPx$. 

The \textit{excess predictive risk} represents the portion of the predictive risk attributable to randomness in the sketching algorithm.  Corollary \ref{cor:predictiverisk} shows that it is equal to
\begin{eqnarray*}
	\mathcal{R} &\equiv& \underbrace{\sigma^2\trace\{\Es[\mP\mP\Tra] - \mPx\}}_{\mathcal{R}_{\mv}} + \underbrace{\|(\Es[\mP\mP\Tra] - \mPx)\mx\vbetao\|_{2}^{2}}_{\mathcal{R}_{\mb}},
\end{eqnarray*}
where the excess predictive variance $\mathcal{R}_{\mv}$ and excess predictive bias $\mathcal{R}_{\mb}$ have interpretation as stated above.

Notice that the bias projector $\mPo$ does not appear in expressions for the total predictive risk.  Therefore, the predictive risk remains unaffected by expected rank preservation and the effects of algorithmic-induced randomness on it are restricted to the deviation of $\mP$ from $\mPx$.  Thus, compared with the total variance and mean squared error for the true parameter, the total predictive risk is less affected by algorithmic-induced randomness.

\begin{corollary}[Total predictive risk conditioned on rank preservation]\label{cor:predictiverisk-rankpreserved}
	For the problem in \eqref{e_s} conditioned on $\rank(\ms\mx) = \rank(\mx)$, the solution $\vtbeta$ has total predictive risk equal to
	\begin{eqnarray*}
		\Risk(\vty, \mx\vbetao) &=& \Risk(\vhy, \mx\vbetao) + \sigma^2\trace\{\Es[\mP\mP\Tra] - \mPx\}.
	\end{eqnarray*}
	Therefore, we additionally have
	\begin{eqnarray*}
		\Risk(\vty, \mx\vbetao) &\ge& \Risk(\vhy, \mx\vbetao).
	\end{eqnarray*}
\end{corollary}

Corollary \ref{cor:predictiverisk-rankpreserved} follows from the following facts when conditioning on $\rank(\ms\mx) = \rank(\mx)$.  First, $\mP\vy$ is an unbiased estimator for $\mx\vbetao$ so that the excess predictive bias $\mathcal{R}_{\mb}$ vanishes.  Second, $\mP\mPx\mP\Tra = \mPx$ so that the excess predictive variance $\mathcal{R}_{\mv}$ is positive semi-definite.

The \textit{excess predictive risk} in this case is given by
\begin{eqnarray*}
	\mathcal{R}' &\equiv& \sigma^2\trace\{\Es[\mP\mP\Tra] - \mPx\} = \mathcal{R}_{\mv},
\end{eqnarray*}
representing the excess predictive variance due to the deviation of $\mynull(\mP)$ from $\mynull(\mPx)$.  Notice that although the bias projector $\mPo$ does not appear in the unconditional total predictive risk in Corollary \ref{cor:predictiverisk}, the predictive risk still decreases when conditioning on rank preservation.  This is because the predictive bias $\Bias(\vty, \mx\vbetao)$ depends only on the deviation of $\range(\mP)$ from $\range(\mPx)$.  Since these are equal when conditioning on $\rank(\ms\mx) = \rank(\mx)$, the predictive bias vanishes in this case.

Notice additionally that although $\range(\mP) = \range(\mPx)$ in this case, we still have $\mynull(\mP) \ne \mynull(\mPx)$ in general.  Therefore, the predictive risk contains excess predictive variance $\mathcal{R}_{\mv}$ arising from the expected deviation of $\mynull(\mP)$ from $\mynull(\mPx)$.

Corollary \ref{cor:predictiverisk-rankpreserved} shows that even when conditioning on sketching schemes that preserve rank so that $\vty$ is an unbiased estimator of $\mx\vbetao$, the total predictive risk of $\vty$ is at least as great as that of $\vhy$.  This is because $\vty$ inherits the predictive variance due to model-induced randomness.  Additionally, it acquires excess predictive variance arising from the perturbation of $\mynull(\mPx)$ under sketching.
\section{Sketching Diagnostics}\label{sec:diagnostics}

In previous sections, we observed that the bias, and hence expected accuracy, of the sketched solution and prediction hinge on rank preservation.  A natural consequence is that the bias projector $\mPo$ proves ideal for use in a sketching diagnostic.  Compared with $\mP \in \mathbb{R}^{n \times n}$, which may be computationally expensive for large $n$, $\mPo \in \mathbb{R}^{p \times p}$ can be computed quickly and inexpensively.  Moreover, if rank is preserved, $\mPo = \mi_{p}$ so that its two-norm condition number $\kappa_{2}(\mPo)$ becomes a simple diagnostic for rank preservation: If $\kappa_{2}(\mPo) = 1$, then the sketching process preserves rank.  Otherwise, it does not.  

%

We illustrate how one can employ $\mPo$ as a sketching diagnostic to aid in the practical design of judicious sketching schemes.  We also show that $\mPo$ can be utilized in selecting a suitable sketching dimension $r$.
To simulate realistic regression data satisfying a Gaussian linear model, we build a linear model based on data from the 2018 American Community Survey (ACS) 1-year Public Use Microdata Sample (PUMS) from the U.S. Census Bureau.  The ACS collects population and housing information on individuals and households across the U.S. to help guide policy-making.  Technical details regarding the ACS PUMS files can be found at \cite{acs2018}.  We employ the ACS PUMS from California as a foundation for realistic survey data from a large and diverse population.

For our initial response $\vy'$, we utilize the gross rent as a percentage of annual household income, and subset for respondents with responses for this variable.  For our initial design $\mx'$, we employ the following economic, language, and household status variables: food stamp program participation, primary household language, limited English proficiency status as a household, multigenerational household status, and citizenship status.  We also employ the following control variables: age, sex, marital status, and education level of the respondent.  We obtain our final design $\mx$ with $n=105{,}142$ respondents and $p=21$ variables after standard recoding for categorical variables and appending a column of ones for the intercept.  
To obtain a Gaussian linear model, we simulate $\vy$ as follows.  We obtain $\vbetao$ by regressing $\vy'$ onto $\mx$ and then setting entries in the resulting estimator corresponding to non-significant variables to zero.  We then obtain $\vy \equiv \mx\vbetao + \veps$, where $\veps$ follows a zero mean multivariate Gaussian distribution with $\sigma^{2} = 10^{-12}$.

We conduct numerical simulations with $\vy$ and $\mx$, and compare each $\vtbeta$ to $\vhbeta$ obtained on the same data.  We compare performance on three sketching schemes: 1) uniform sampling with replacement (UNIF), 2) unweighted leverage score sampling with replacement (LEV) \cite{MMY14, MMY15}, and 3) random projections with a matrix whose entries are standard Gaussian random variables (NORM).  To illustrate how rank preservation varies with $r$, we perform simulations over a range of sketching dimensions.  These range from $r=20 < 21 = p$, so that all simulations perform poorly, to $r=100$, where most simulations perform well.  We run $100$ replicates of each scenario.


\begin{figure}[t]
	\centering
	\begin{subfigure}{.5\textwidth}
		\centering
		\includegraphics[width=\linewidth]{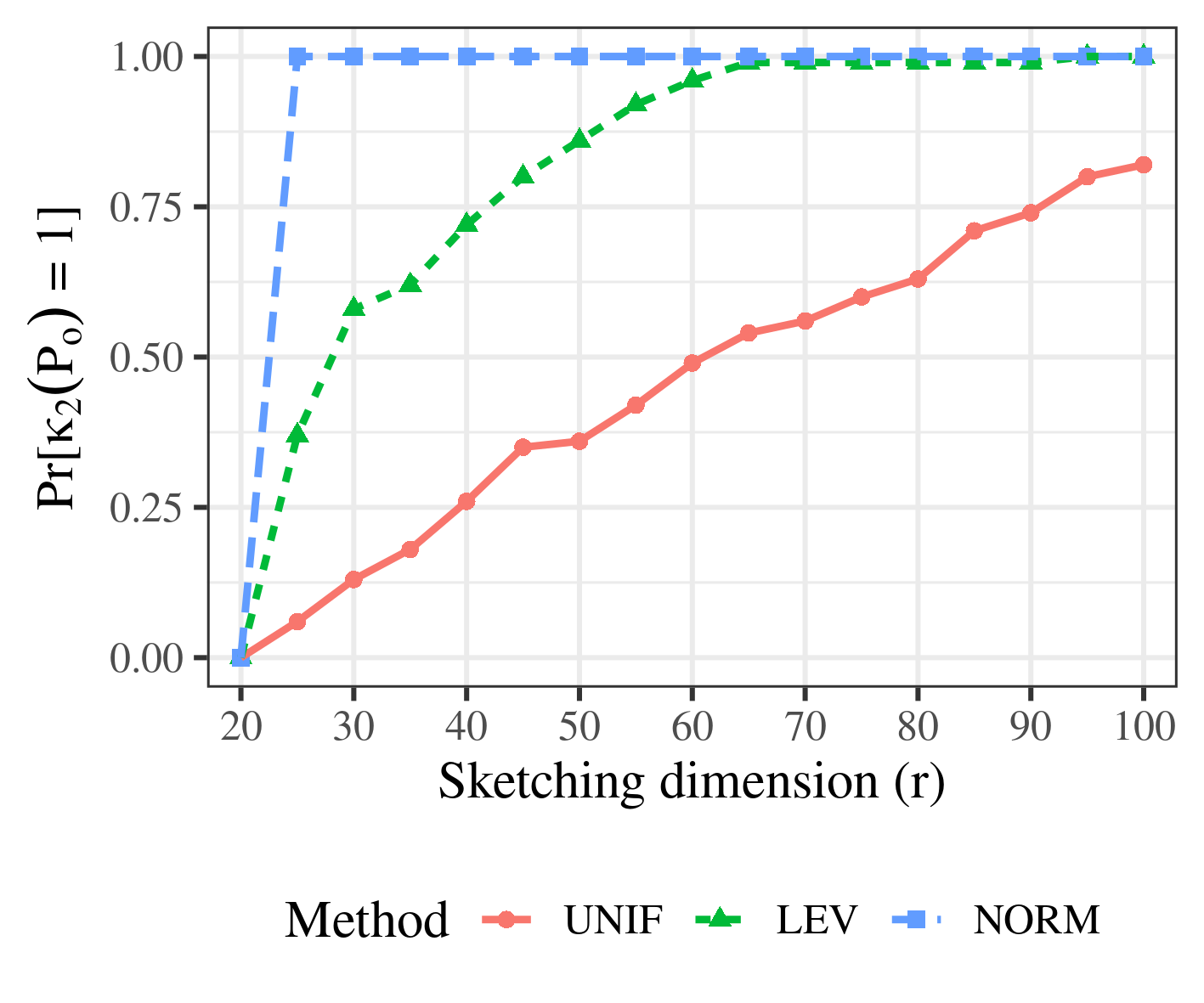}
		\caption{Estimated probability of rank preservation}
		\label{fig:kappa}
	\end{subfigure}%
	\begin{subfigure}{.5\textwidth}
		\centering
		\includegraphics[width=\linewidth]{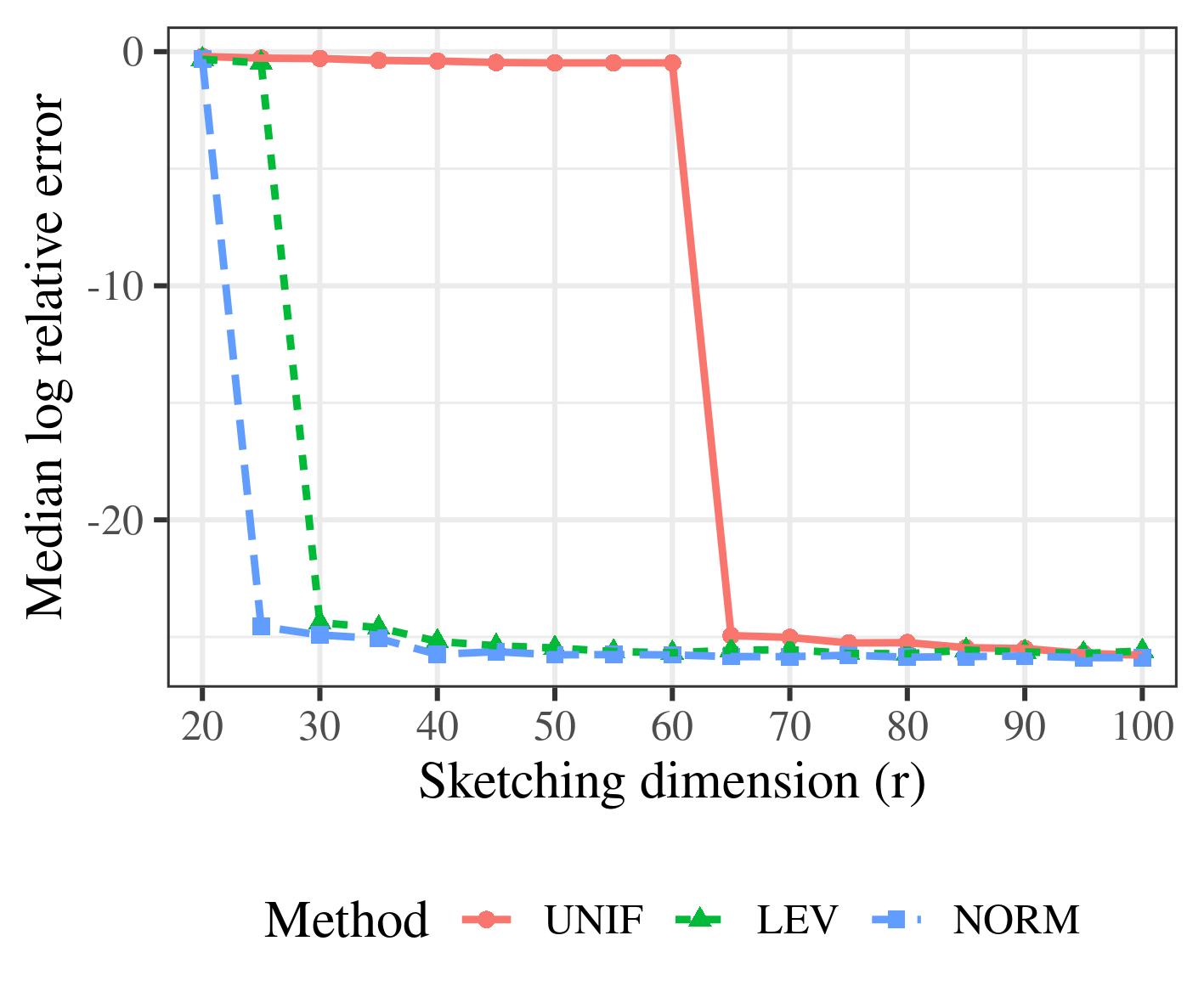}
		\caption{Median log relative error}
		\label{fig:relerror}
	\end{subfigure}
	\caption{\textit{Simulation results illustrate the pattern between rank preservation (a) and median log relative error of $\vtbeta$ with respect to $\vhbeta$ (b) as a function of sketching method and dimension.}}
	\label{fig:sims}
\end{figure}

Figure \ref{fig:kappa} depicts Pr$[\kappa_{2}(\mPo) = 1]$, the estimated probability of rank preservation, over the $100$ replicates for each scenario.  We observe that the $r$ at Pr$[\kappa_{2}(\mPo) = 1] > 0.50$ corresponds to the $r$ where the relative error transitions from high to low in Figure \ref{fig:relerror}.  NORM and LEV achieve Pr$[\kappa_{2}(\mPo) = 1] > 0.5$ at $r=25$ and $r=30$, respectively, and their relative errors likewise drop then.  UNIF achieves Pr$[\kappa_{2}(\mPo) = 1] > 0.5$ at $r=65$ so it transitions to low relative error at $r=65$.

Figure \ref{fig:sims} illustrates that since $\kappa_{2}(\mPo) = 1$ correlates with low relative error, it can provide an inexpensive diagnostic for candidate sketching matrices.  Figure \ref{fig:sims} also shows that given a class of sketching matrices, one can employ Pr[$\kappa_{2}(\mPo) = 1$] in selecting an appropriate $r$.  For example, in this illustrative problem, the numerical results shown in Figure \ref{fig:sims} would suggest selecting $r=25$ if employing Gaussian sketching.  This may be useful in solving large iterative linear systems where it may be impractical to hand-select a sketching matrix at each iteration.

\section{Discussion}\label{s_discuss}

We presented a projector-based approach for sketched linear regression and analyzed the combined uncertainties on the sketched solution $\vtbeta$ from both statistical noise in the model and randomness from the sketching algorithm.  Our results show that the total expectation and variance of $\vtbeta$ are
governed by the spatial geometry of the sketching process, rather than by structural properties of 
specific sketching matrices. Surprisingly, the condition number $\kappa_2(\mx)$ with respect to (left) inversion has far less impact on the statistical measures than it has on the numerical errors. 

Our results demonstrate the usefulness of a projector-based approach in enabling expressions for quantifying the total and excess uncertainties that hold generally for \textit{all} sketching schemes.  A projector-based approach also enables insights and interpretations on how the sketching process affects the solution and other key statistical quantities.  Finally, our numerical experiments illustrate the practicality of the bias projector $\mPo$ as a computationally inexpensive and effective sketching diagnostic under a Gaussian linear model.

%
%
%
%
%

\bibliographystyle{agsm}
\bibliography{randLS}
\end{document}